%% file: main.tex
\documentclass{article}

\usepackage[final]{corl_2026} % Uncomment for the camera-ready ``final'' version.
\usepackage{graphicx}
\usepackage{multirow}
\usepackage{makecell}
\usepackage{appendix}

\input{preamble}

\title{\spaceskip=0.20em plus 0.1em minus 0.1em ReactiveBFM: Reactive Closed-Loop Motion Planning Towards Universal Humanoid Whole-Body Control}
\newcommand{\blfootnote}[1]{%
  \begingroup
  \renewcommand\thefootnote{}\footnote{#1}%
  \addtocounter{footnote}{-1}%
  \endgroup
}

\author{
  \textbf{Xiao Chen$^{1, 2}$ \quad Weishuai Zeng$^{2*}$ \quad Xiaojie Niu$^{2*}$ \quad Zirui Wang$^{2*}$ \quad Jianan Li$^{1*}$} \\[2pt]
  \textbf{Huayi Wang$^{2}$ \quad Furui Xu$^{2}$ \quad Jiahe Chen$^{2}$ \quad Weixiang Zhong$^{2}$ \quad Lihe Ding$^{1}$} \\[2pt]
  \textbf{Kailin Li$^{2}$ \quad Jiangmiao Pang$^{2}$ \quad Tai Wang$^{2}$ \quad Tianfan Xue$^{1,2\dagger}$ \quad Jingbo Wang$^{2\dagger}$} \\[5pt]
  \textnormal{$^1$The Chinese University of Hong Kong \quad $^2$Shanghai AI Laboratory} \\[5pt]
  \textbf{Project Website}: \href{https://xiao-chen.tech/reactivebfm}{\texttt{xiao-chen.tech/reactivebfm}}
}

\begin{document}

\begin{center}
    \maketitle
    \vspace{-30pt}
    \blfootnote{$^*$Core contributors (listed in random order). $^\dagger$Corresponding authors.}
    \includegraphics[width=1.0\textwidth]{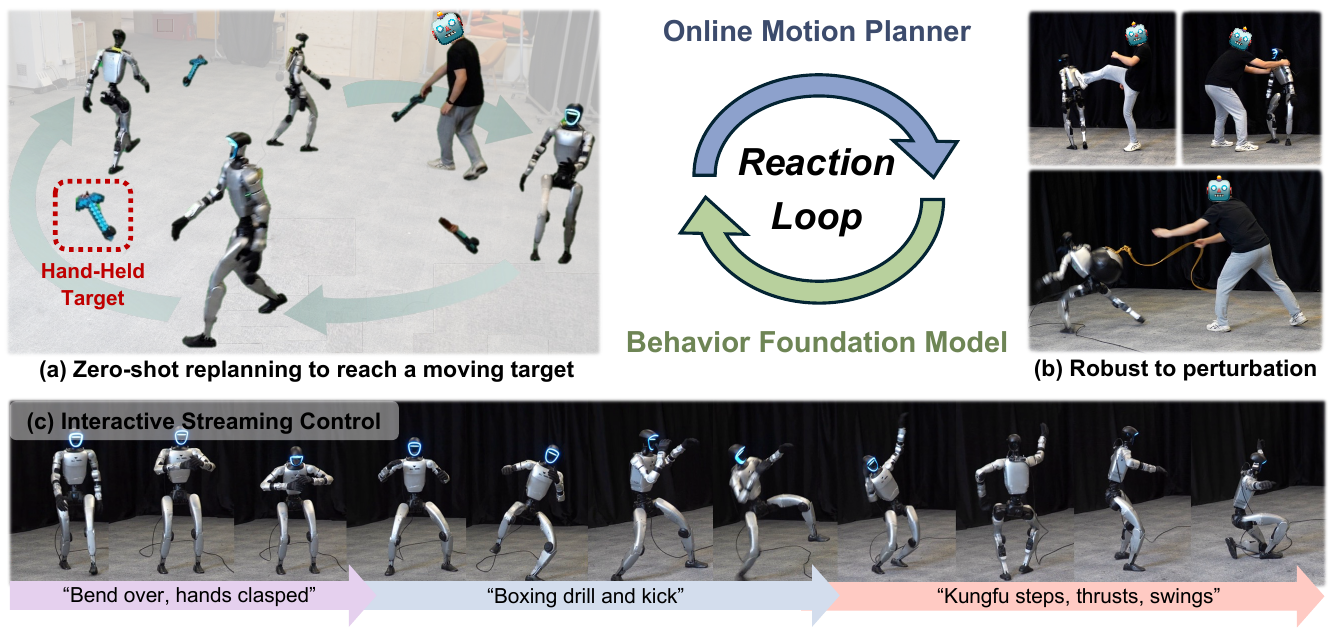}
    \vspace{-20pt}
    \captionof{figure}{
    We introduce \textbf{\ours}, a closed-loop framework integrating a behavior foundation model with a reactive whole-body motion planner.
    Guided by proprioceptive feedback, text, and target positions, \ours enables robust text-conditioned control and seamless zero-shot replanning to reach moving targets.
    }
    \label{fig:teaser}

\end{center}

\input{sections/0_abstract}

\input{sections/1_introduction}

\input{sections/2_related_work}
\input{sections/3_method}

\input{sections/4_experiments}

\input{sections/5_conclusion}

%===============================================================================

% \clearpage
% The acknowledgments are automatically included only in the final and preprint versions of the paper.

\section*{Acknowledgments}
\vspace{-3pt}
% This work is supported in part by the Centre for Perceptual and Interactive Intelligence (CPII) Ltd., a CUHK-led InnoCentre under the InnoHK initiative of the Innovation and Technology Commission of the Hong Kong Special Administrative Region Government, the National Key R\&D Program of China (2022ZD0160201), and Shanghai Artificial Intelligence Laboratory.~\xiao{TODO}
% We thank all anonymous reviewers for their constructive feedback and valuable suggestions.
This study was supported by Shanghai Artificial Intelligence Laboratory and CUHK-CUHK(SZ)-GDSTC Joint Collaboration Fund No. 2025A0505000053. We sincerely thank Yuxi Wei, Ke Fan, Tao Huang, Junli Ren, and Weiji Xie for their constructive advice and insightful discussions. We also thank Kinetix AI for their support with the HTC VIVE Ultimate Trackers.

\textbf{Author contributions.} Xiao Chen led and implemented all components of this project. Weishuai Zeng, Zirui Wang, Weixiang Zhong, and Jiahe Chen assisted in training and adapting the BFM foundation model, developed the motion data infrastructure, and participated in discussions regarding the system architecture and optimization strategies. Xiaojie Niu participated in all real-world deployments and designed the communication and data processing pipeline of global localization. Jianan Li and Lihe Ding participated in the architecture design and training formulation of the reactive planner. Huayi Wang and Furui Xu assisted in the data collection and post-processing for the PhysHSI-Reach dataset. Jingbo Wang, Tianfan Xue, Tai Wang, Jiangmiao Pang, and Kailin Li jointly supervised and co-led the research project.
% }

% \clearpage
%===============================================================================

\bibliography{references}  % .bib

\input{sections/6_appendix}

\end{document}

%% file: preamble.tex
\usepackage{multirow}
\usepackage{multicol}
\usepackage{graphicx}
\usepackage{xspace}
\usepackage{xcolor}
\usepackage{caption}
\usepackage{wrapfig}
\usepackage{bbding}  % For \Checkmark
\usepackage{pifont}  % For \XSolidBrush

\usepackage{booktabs}
\usepackage{float}
\usepackage{amsmath}  % 提供了高级数学公式支持
\usepackage{amssymb}  % 提供了额外的数学符号
\usepackage{bm}
\usepackage{dsfont}

\usepackage{natbib}
\PassOptionsToPackage{numbers,sort&compress}{natbib}

\usepackage{hyperref}
\usepackage[capitalise, nameinlink]{cleveref}
\newcommand{\paragraphbegin}[1]{\vspace{0.02in}\noindent\textbf{#1}}

\newcommand{\ours}[0]{ReactiveBFM\xspace}

\usepackage{colortbl}
\definecolor{ourcolor}{HTML}{99e0eb}
\definecolor{ourblue}{HTML}{27a2c3}

\definecolor{tablecolor}{HTML}{ccf2f5} 

\definecolor{tablecolor2}{HTML}{ffcdb4}
\definecolor{citecolor}{HTML}{fe7b5b}
\definecolor{grey}{rgb}{0.9, 0.9, 0.9}
\usepackage{amssymb}

\usepackage{listings}
\definecolor{gred}{rgb}{0.859,0.267,0.216}
\definecolor{ggreen}{rgb}{0.059,0.616,0.345}

\definecolor{deepblue}{HTML}{27a2c3}

\definecolor{deepred}{HTML}{fe7b5b}

\usepackage[font=small,labelfont=bf]{caption}

\usepackage[font=footnotesize,labelfont=bf]{caption}

\definecolor{citecolor}{HTML}{faa700} %3953A4
\definecolor{lblue}{HTML}{ffb114} %3953A4 0071bc
\definecolor{ogreen}{HTML}{2E7D32}
\definecolor{bred}{HTML}{BF360C}
\definecolor{newbrown}{HTML}{795548}

\hypersetup{
    colorlinks=true,
    linkcolor=blue,
    filecolor=magenta,      
    urlcolor=blue,
    citecolor=blue,
}

\usepackage{multicol}
\usepackage{multirow}
\usepackage{colortbl}
\usepackage{booktabs}   % for rule
\usepackage{bbding} % markers
\usepackage{graphicx}
 
\usepackage{enumitem}

\usepackage{bbding}

\usepackage{pifont}
\usepackage{xfrac}
\usepackage{tikz}

\usepackage[para,online,flushleft]{threeparttable}

%% file: sections/0_abstract.tex
% The purpose of this document is to provide both the basic paper template and submission guidelines. Abstracts should be a single paragraph, between 4--6 sentences long, ideally. Gross violations will trigger corrections at the camera-ready phase.
\vspace{-8pt}
\begin{abstract}
    % Developing Behavior Foundation Models (BFMs) from human motion priors has emerged as a promising pathway toward general-purpose humanoid control.
    % However, these models typically operate as open-loop execution pipelines bound to pre-existing reference trajectories, inherently lacking the system-level closed-loop mechanisms needed to synthesize motions reactively in response to dynamic environments and actual physical states.
    % To bridge this gap, we propose\textbf{~\ours}, a novel online closed-loop control framework comprising a reactive whole-body motion planner and a universal tracking controller.~\xiao{technical motivation of planner and controller}
    % Specifically, our motion planner interactively generates whole-body kinematics in real time by conditioning on language instructions, dynamic spatial targets, and the robot's actual historical proprioception.
    % This real-time proprioception feedback is continuously provided by our robust universal tracker, which zero-shot executes the predicted diverse motions in the global frame. By tightly coupling these two modules into a reactive replanning loop, our system entirely eliminates the need for pre-existing reference motions.
    % Extensive real-world experiments highlight our system's zero-shot capabilities in dynamic target reaching and expressive motion generation under severe physical disturbances. These results demonstrate that our closed-loop synergy achieves exceptional robustness and versatility.
    While current Behavior Foundation Models (BFMs) provide robust control priors for humanoids, they only execute pre-defined reference motions. As a result, they are vulnerable to environmental shifts and incapable of reactive whole-body coordination. 
    Naively cascading them with generative motion planners fails to achieve true reactivity, as inevitable tracking discrepancies induce fatal cumulative exposure bias.
    To bridge this gap, we propose~\textbf{\ours}, a real-time closed-loop planning-control framework. 
    At its core, we effectively mitigate exposure bias via a scheduled prefix sampling curriculum, forcing the generative planner to actively learn error-recovery behaviors from imperfect physical states rather than ground-truth trajectories.
    Systematically, to reconcile the severe latency mismatch between auto-regressive planning and high-frequency tracking, we introduce an asynchronous replanning mechanism. Combined with trajectory chunking to temporally ensemble spatial references, our system guarantees spatio-temporally fluid execution without physical jitter.
    Deployed on the Unitree G1 humanoid, \ours demonstrates unprecedented physical agility across a vast repertoire of text-conditioned closed-loop motions. Notably,~\ours achieves zero-shot moving target reaching, showcasing intricate whole-body coordination and on-the-fly replanning. 
    In sim-to-sim benchmarking under severe perturbations, \ours achieves a 93.1\% success rate, significantly outperforming cascaded open-loop baselines by 28.6\%. 
    Our code is available at \href{https://github.com/zjwzcx/ReactiveBFM}{https://github.com/zjwzcx/ReactiveBFM}.

    % Systematically, to reconcile the latency mismatch between auto-regressive planning and high-frequency tracking, we design an asynchronous replanning architecture. By integrating trajectory chunking to temporally ensemble spatial references, \ours guarantees spatio-temporally fluid hardware execution without physical jitter.
    % During real-world deployment, we design asynchronous replanning architecture to overcome latency mismatch. Further applying trajectory chunking to temporally ensemble spatial references, \ours guarantees spatio-temporally fluid hardware execution without physical jitter.
    % While this curriculum grants algorithmic robustness, deploying such a closed-loop generative system in the real world exposes a severe latency bottleneck.
    % To ensure closed-loop stability, we introduce scheduled AR prefix sampling to eliminate cumulative prediction errors, alongside an event-driven asynchronous architecture with temporal ensembling to resolve latency and spatio-temporal conflicts.
    % Deployed on the Unitree G1 humanoid, \ours demonstrates unprecedented physical agility and error-recovery. Notably, via a mathematically elegant ego-centric coordinate reset, the system achieves zero-shot long-horizon dynamic target reaching without task-specific data, alongside seamless task-switching under streaming multi-modal instructions.

    % Furthermore, rigorous streaming interactive control and robustness evaluations explicitly highlight the framework's exceptional conditional stability and error-recovery capabilities in unpredictable physical environments.

\end{abstract}

%% file: sections/1_introduction.tex
\section{Introduction}
\vspace{-5pt}
% Developing general-purpose humanoid robots demands the mastery of a vast repertoire of diverse motor skills while navigating unpredictable real-world dynamics. When deployed in unstructured environments, humanoids are inevitably subjected to unforeseen physical interactions and evolving task constraints. 
% Consequently, maintaining physical viability in such dynamic settings requires more than just executing isolated skills. A robust autonomous system must seamlessly integrate high-level, context-aware motion generation with low-level, real-time reactive control.
% % Consequently, a robust autonomous system must not only generate complex, context-aware whole-body motions but also continuously provide reactive stabilization against unexpected environment changes to maintain physical viability.
Developing general-purpose humanoid robots demands the mastery of a vast repertoire of diverse motor skills while navigating unpredictable real-world dynamics. When deployed in unstructured environments, humanoids are inevitably subjected to unforeseen physical interactions and evolving task constraints. Thus, a robust autonomous system must consequently function as a reactive planner capable of continuously generating and adapting context-aware whole-body trajectories on the fly.

While recent advancements in universal trackers~\citep{yin2025unitracker,ze2025twist2,luo2025sonic} and Behavior Foundation Models (BFMs)~\citep{zeng2025behavior,li2025bfm} provide robust low-level control priors, they strictly demand precise trajectory inputs. Simply cascading them with existing motion generators fails to achieve true reactivity. Current generators predominantly perform offline synthesis~\citep{wei2025unveiling,li2025learning,kalaria2025dreamcontrol} or sparse keyframe in-betweening~\citep{li2025genmo,luo2025sonic}, rigidly enforcing an open-loop paradigm. Completely blind to real-time environmental shifts and state transitions, these approaches cause the robot to rapidly accumulate severe physical drift.
\textbf{To achieve genuine reactivity, the generative planner must operate in an online, closed-loop manner.} 
However, this architectural shift fatally collides with cumulative exposure bias. Typical planners~\citep{chen2024taming,zhao2025dartcontrol,tevetclosd}, accustomed to perfect historical inputs via teacher-forcing, are completely unequipped to handle inevitable tracking discrepancies. Once pushed out of their training distribution, they fail to recover, causing execution errors to compound recursively. % and ultimately crash the system.

To bridge these critical gaps, we propose \textbf{\ours}, a closed-loop planning-control framework that fundamentally resolves \textit{cumulative exposure bias} in reactive humanoid control. 
Our core solution is a scheduled prefix sampling curriculum for planner training. Instead of relying on perfect ground-truth prefixes, the model actively learns error-recovery behaviors to survive imperfect physical states. 
Second, a reactive system must seamlessly handle mid-execution command shifts. We integrate condition dropout and temporal regularization to prevent jerky transitions, ensuring spatio-temporal smoothness during runtime instruction updates.
Finally, redundant kinematic features inherently accelerate physical drift. We strip away this over-parameterization, utilizing a compact representation to strictly bound the error space.
% We also reinforce this reactive stability by bounding the physical drift space through a compact kinematic representation.
% At the low level, these continuous spatial targets are seamlessly tracked by the BFM trained on 64 GPUs for about 10 days using over 560 hours of diverse motion data, delivering unprecedented global tracking precision and zero-shot sim-to-real transfer capabilities.

Beyond algorithmic robustness, real-world deployment faces a strict \textit{latency bottleneck}, as auto-regressive generation inherently operates at a lower frequency than motion tracking. 
We resolve this frequency mismatch via asynchronous replanning. 
To eliminate the physical jitter caused by overlapping asynchronous predictions, we apply trajectory chunking~\citep{zhao2023learning,fu2024mobile} to temporally ensemble the spatial references, ensuring fluid hardware execution.

We extensively validate \ours through rigorous sim-to-sim benchmarking and real-world deployments on the Unitree G1 humanoid. 
\ours achieves a 90\% success rate and sustains over 40 seconds of continuous execution in zero-shot dynamic target reaching, successfully generalizing without any dynamic-target training data.
Furthermore, our robust streaming interactive control enables seamless transitions between long-horizon maneuvers under continuous text commands.
Sim-to-sim benchmarking reveals our closed-loop architecture achieves a 93.1\% task success rate and reduces the fall rate to merely 2.0\% under severe perturbations. It significantly outperforms cascaded open-loop baselines, such as TextOp~\citep{xie2026textop} combined with SONIC~\citep{luo2025sonic}, by 28.6\%.

%% file: sections/2_related_work.tex
\vspace{-7pt}
\section{Related Work}
\vspace{-4pt}
\subsection{Whole-Body Motion Generation}
    Diffusion- and autoregressive-based generative models, first demonstrated at scale in image synthesis~\citep{ho2020denoising,nichol2022glide,ramesh2022hierarchical,rombach2022high}, have been widely adopted for kinematic motion synthesis~\citep{tevet2023human,zhang2023generating,jiang2023motiongpt,chen2024taming}.
    Recent efforts further push motion generators towards autoregressive streaming, real-time interaction, and scaled-up training~\citep{harvey2020robust,lu2025scamo,xiao2025motionstreamer,zhao2025dartcontrol,rempe2026kimodo,xie2026textop}.
    These generators are typically trained on open-source human MoCap datasets~\citep{mahmood2019amass,guo2022generating,punnakkal2021babel}, on large-scale human motion corpora~\citep{lin2023motion,fan2025go}, or on robot-space corpora retargeted from such sources~\citep{wei2025unveiling}.
    However, without rigorous physical constraints, these datasets frequently exhibit physically implausible artifacts, such as foot sliding and self-penetration. Directly injecting these uncorrected priors into a closed-loop framework inevitably triggers severe sim-to-real gaps.

\subsection{Universal Planning-Control Systems}
    Early reinforcement learning controllers master individual agile skills, such as parkour~\citep{zhuanghumanoid,zhuang2026deep}, Kungfu~\citep{xie2025kungfubot}, and fall recovery~\citep{huang2025learning}, but rely on handcrafted rewards and task-specific training that hinder generalization.
    Recent Behavior Foundation Models (BFMs) and universal tracking controllers~\citep{luo2023perpetual,cheng2024expressive,ji2024exbody2,chen2025gmt,ze2025twist2,luo2025sonic,zeng2025behavior,li2025bfm,huang2025towards,yin2025unitracker} have emerged as robust low-level execution engines. By scaling up reinforcement learning across massive, diverse datasets~\citep{mahmood2019amass,mason2022real}, these policies track unseen kinematic references with high precision without task-specific adaptation.

    Despite their impressive tracking fidelity, integrating these models into autonomous systems exposes a critical architectural flaw: the reliance on open-loop execution. Frameworks like Mimic2DM~\citep{li2025learning}, DreamControl~\citep{kalaria2025dreamcontrol}, and Imagine2Real~\citep{chen2026imagine2real} restrict planning to offline reference extraction.
    Even when utilizing text-conditioned planners like TextOp~\citep{xie2026textop}, the generated trajectories remain fundamentally blind to subsequent environmental changes.
    More restrictively, SONIC~\citep{luo2025sonic} reduces motion planning to keyframe in-betweening, forcing the robot to rigidly interpolate between predefined poses and severely crippling its dynamic reactivity.
    Similarly, VisualMimic~\citep{yin2025visualmimic} restricts its generation space to a sparse 6-keypoint interface, which is inadequate for dense whole-body tracking.
    While methods such as CLoSD~\citep{tevetclosd} and DartControl~\citep{zhao2025dartcontrol} introduced recursive closed-loop architectures, their efficacy remains confined to simulation, leaving real-world deployment unresolved.

%% file: sections/3_method.tex
\section{\ours: A Closed-Loop Planning-Control System}
\label{method}
\begin{figure}[!t]
        \centering
        \includegraphics[width=1\linewidth, height=0.5\linewidth]{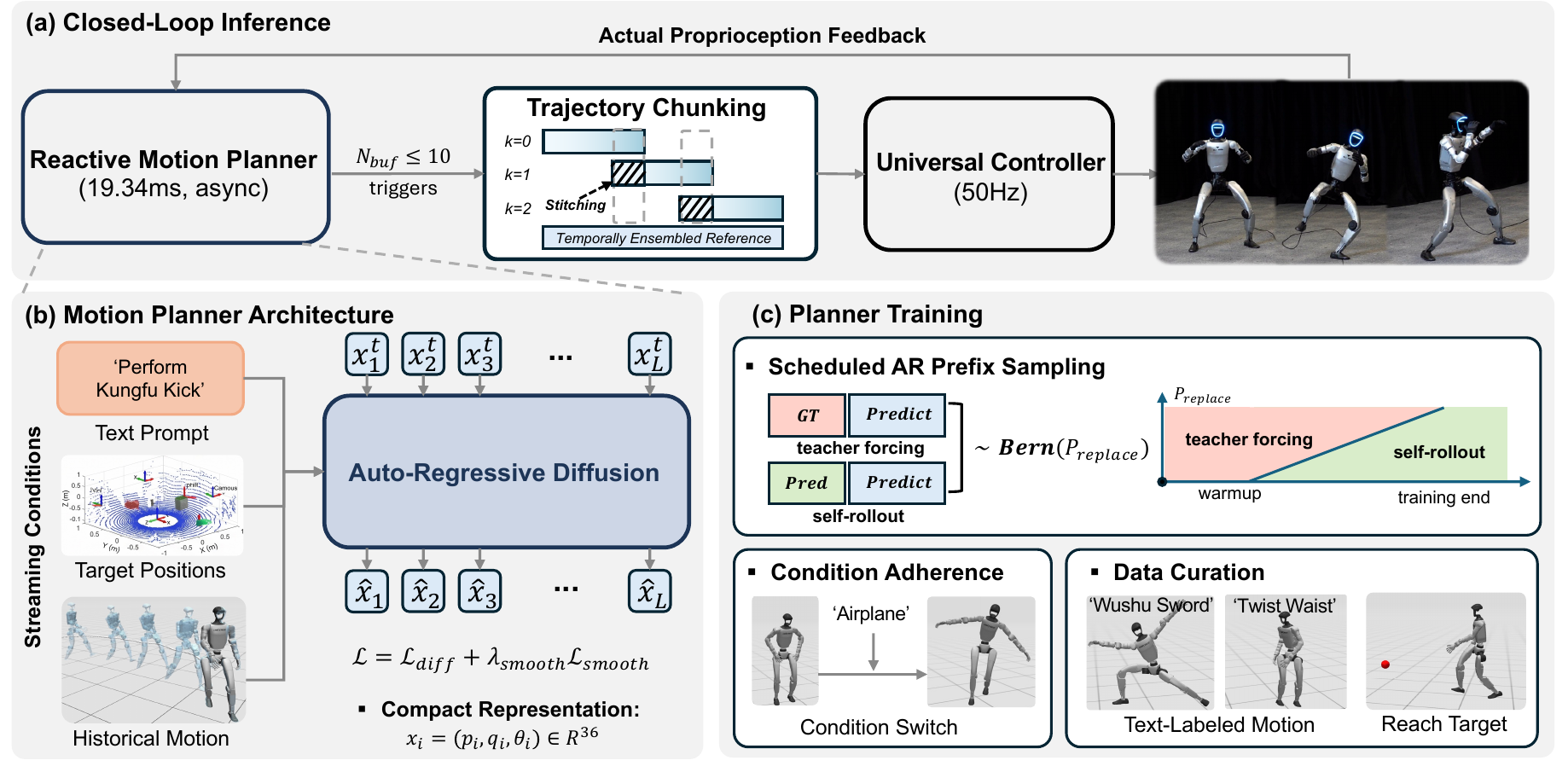}
        \caption{\textbf{Overview of \ours.} (a) Asynchronous closed-loop inference coupled with a universal controller via trajectory chunking and proprioceptive feedback. (b) Architecture of our reactive motion planner. It generates smooth robot trajectories from multi-modal streaming conditions. (c) Core training strategies including a scheduled prefix sampling curriculum and condition adherence to mitigate exposure bias.}
        \label{fig:method}
        \vspace{-18pt}
    \end{figure}

    % In this section, we present our closed-loop planning-control framework~\ours, as illustrated in Fig.~\ref{fig:method}. We first detail our reactive whole-body motion planner in Sec.~\ref{method:gen} and our BFM controller in Sec.~\ref{method:bfm}. Finally, Sec.~\ref{method:system} elaborates on our closed-loop architecture and real-world deployment.
    In this section, we present our closed-loop planning-control framework~\ours, as illustrated in Fig.~\ref{fig:method}. We first detail the training strategies of the reactive whole-body motion planner in Sec.~\ref{method:gen}. Next, Sec.~\ref{method:system} elaborates on our closed-loop architecture and real-world deployment. Finally, we provide an overview of the key implementation details of our BFM in Sec.~\ref{method:bfm}.

\subsection{Reactive Whole-Body Motion Planner}
\label{method:gen}
    We introduce an auto-regressive (AR) motion diffusion model as a reactive, closed-loop planner.
    The primary bottleneck in deploying AR planners is \textbf{cumulative exposure bias}. 
    The execution errors compound recursively, shifting the model out of its training distribution. 
    To break this cycle and ensure robust execution, we propose the following designs.
    
    % To bridge the gap between open-world generation and closed-loop execution, we propose four key designs to effectively learn a robust planner: (a) a \textbf{compact kinematic representation} to eliminate physical inconsistencies, (b) \textbf{scheduled auto-regressive prefix sampling} to stabilize long-horizon rollouts, (c) \textbf{temporal consistency loss} to ensure replanning smoothness, and (d) \textbf{high-quality data curation} for physically grounded learning.

    \paragraphbegin{Auto-Regressive Formulation and Compact Representation.}
    We formulate online reactive whole-body planning as an auto-regressive motion generation problem~\citep{tevet2023human,tevetclosd,chen2024taming}. 
    % Unlike traditional in-betweening paradigms~\citep{harvey2020robust,cohan2024flexible} that rely on rigid keyframe priors,
    % The AR approach recursively synthesizes future trajectories conditioned on continuous multi-modal feedback, intrinsically aligning with predictive motion control.
    Conditioned on language instructions, target positions, and historical proprioception, our AR model synthesizes feasible reference trajectories on the fly in real time. 
    
    Traditional representations~\citep{guo2022generating,fan2025go,xiao2025motionstreamer} often adopt densely augmented kinematic representations containing contact states or global velocities. While geometrically descriptive, this over-parameterization burdens the generative planner and causes kinematic misalignment.
    % and frequently causes misalignments that destabilize closed-loop execution. 
    To resolve this, we define a minimal yet kinematically complete 36-dim representation:
    \begin{equation}
        \mathbf{x}_i = [\mathbf{p}_i, \mathbf{q}_i, \boldsymbol{\theta}_i] \in \mathbb{R}^{36},
    \end{equation}
    where $\mathbf{p}_i \in \mathbb{R}^3$ and $\mathbf{q}_i \in \mathbb{R}^4$ denote the root's global translation and quaternion rotation, respectively, and $\boldsymbol{\theta}_i \in \mathbb{R}^{29}$ encapsulates the joint positions corresponding to the 29 DoFs of Unitree G1.
    
    Operating over overlapping windows of this continuous state space $\mathbf{X}_{1:L} = [\mathbf{x}_1, \dots, \mathbf{x}_L]$, we instantiate our planner using an Auto-Regressive Motion Diffusion Model (AR-MDM). For a local motion chunk $\mathbf{X}_0$, the generative model $\mathbf{G}$ predicts the clean chunk $\hat{\mathbf{X}}_0 = \mathbf{G}(\mathbf{X}_\tau, \mathbf{c}, \tau)$ from the noisy signal $\mathbf{X}_\tau$ at diffusion timestep $\tau \in [1, T]$. The condition $\mathbf{c}$ incorporates context such as text instructions, target positions, and a prefix trajectory from the preceding window. By directly defining the generative space along the robot's proprioceptive interfaces, the main objective is streamlined to a standard Mean Squared Error (MSE) loss in the motion space:
    \begin{equation}
        \mathcal{L}_{diff} = \mathbb{E}_{\mathbf{X}_0 \sim p(\mathbf{X}_0|\mathbf{c}), \tau \sim [1, T]} \left[ \| \mathbf{X}_0 - \hat{\mathbf{X}}_0 \|_2^2 \right].
    \end{equation}

    \paragraphbegin{Scheduled Auto-Regressive Prefix Sampling.}
    A fundamental challenge in deploying generative planners within a closed-loop paradigm is mitigating the compounding errors caused by open-world disturbances and imperfect downstream tracking. Traditional kinematic motion generators~\citep{tevet2023human,tevetclosd} heavily rely on teacher forcing during training, conditioning exclusively on perfect ground-truth prefix trajectories. However, this creates a severe exposure bias. When deployed, any minor discrepancy between the generated reference and the controller's actual execution forces the planner out of its training distribution, often leading to cascading failures.

    To ensure robust closed-loop execution, we propose a curriculum learning paradigm. During the initial training phase, we employ teacher forcing to rapidly warm up the diffusion model. As training progresses, we introduce scheduled sampling by linearly decaying the probability of using ground-truth prefixes. Instead, the model transitions into a \textit{self-rollout} mode, where it autoregressively conditions on its own previously generated predictions over an extended horizon. To further close the sim-to-real gap, we inject Gaussian noise into the prefix representations as domain randomization. This forces the generative planner to learn error-recovery behaviors, ensuring it can stably output physically plausible trajectories even when the current robot state diverges from ideal conditions.

    \paragraphbegin{Replanning Smoothness and Condition Adherence.}
    Beyond prefix stability, a practical closed-loop system must maintain spatiotemporal smoothness and dynamically respond to streaming multi-modal instructions. 
    First, standard MDMs typically optimize a point-wise MSE loss, which is agnostic to temporal dynamics. This often yields jerky motions that fail to smoothly connect with the robot's current state. To explicitly enforce kinematic consistency, we augment the training objective with a temporal consistency loss, including first-order (velocity) and second-order (acceleration). The temporal regularization penalizes non-smooth transitions, ensuring holistic trajectory smoothness and zero-order continuity at the prefix-generation boundary.
    Second, to prevent the planner from collapsing into a single conditioning modality, we apply independent dropout to the text instructions and target positions during training. As a result, our planner captures unconditioned motion prior.
    Also, we simulate cross-motion transitions to handle streaming interactive control where users may issue new commands mid-execution.
    Consequently, when reacting to unpredictable conditional shifts, the model relies on this generalized prior to synthesize stable transitions. This fundamentally prevents out-of-distribution tracking errors from escalating into cumulative exposure bias.

    \paragraphbegin{High-Quality Data Curation.}
    Raw kinematic datasets often contain physically infeasible artifacts, such as foot sliding and self-penetration, which rapidly destabilize closed-loop systems through error accumulation. To prevent our planner from proposing unreachable states, we process diverse text-motion corpora including 100STYLE~\citep{mason2022real}, AMASS~\citep{mahmood2019amass}-HumanML3D~\citep{guo2022generating}) and Kungfu~\citep{lin2023motion} through a rigorous physical curation pipeline. First, we enforce kinematic canonicalization and explicit physical corrections via simulation~\citep{coumans2016pybullet} to strictly eliminate dynamically invalid motions. 
    In addition, we synthesize 10,000 goal-conditioned reaching trajectories using a pre-trained PhysHSI~\citep{wang2025physhsi} policy.
    Together, these two data streams yield 37.14 hours of dynamically verified motion data. Ultimately, this corpus equips the planner to translate high-level text intents and dynamic spatial targets into safe, hardware-executable trajectories.

    \paragraphbegin{Planner Architecture.} 
    To ensure the real-time inference frequency, our planner adopts a lightweight architecture employing an 8-layer transformer decoder~\citep{vaswani2017attention}, following~\citep{tevetclosd,chen2024taming}. Our transformer-based planner is optimized via a standard diffusion objective~\citep{ho2020denoising}.

    \paragraphbegin{Planner Optimization.}
    To ensure physical continuity between past and future states, we formulate the trajectory generation as a conditional motion inpainting task~\citep{lugmayr2022repaint,janner2022planning}. Specifically, the primary motion stream takes the temporal concatenation of the observed motion prefix and the noisy future motion, processing them jointly through bidirectional self-attention to capture full sequence context. Concurrently, the diffusion time step embedding, and the text instruction tokens are concatenated to form a unified memory stream. This conditioning sequence is subsequently injected into the motion stream via cross-attention layers, allowing the denoising process of the future frames to dynamically query both the semantic intent and the temporal diffusion schedule.

\subsection{Closed-Loop System \& Real-World Deployment}
\label{method:system}
    % While our planner and controller share a unified kinematic interface, directly cascading them into a naive serial pipeline is impractical for real-world operations. 
    % Deploying this closed-loop system introduces three core challenges: latency bottlenecks, spatio-temporal consistency, and stable-to-dynamic generalization.
    While our planner provides robust priors, a naive planner-controller cascade often fails under real-world physical disturbances. To enable robust deployment, our closed-loop system resolves three critical challenges: inference latency, replanning consistency, and stable-to-dynamic generalization.
    
    \paragraphbegin{Asynchronous Replanning and Acceleration.} 
    To prevent the planner from blocking the 50 Hz control loop, we implement an event-driven asynchronous execution mechanism. Whenever the unexecuted reference frames in the control buffer drop below a safe margin $N_{buf}=10$ frames, the system asynchronously extracts the latest historical proprioception and triggers a non-blocking planner thread.
    To minimize the inherent inference and communication latency, we aggressively optimize both networks using TensorRT compilation, reducing the planner's latency to 19.3 ms and the controller's to 5.9 ms. Crucially, we strictly isolate the controller process with real-time CPU scheduling priority during deployment. This ensures the 50 Hz control loop is never preempted by the planner's inference, maintaining stringent real-time responsiveness without compute starvation.
    
    \paragraphbegin{Temporal Ensembling and Spatial Alignment.} 
    Asynchronous replanning inherently generates overlapping reference trajectories, which can induce high-frequency oscillations during deployment. To mitigate this, we employ temporal ensembling~\citep{zhao2023learning,fu2024mobile} via trajectory chunking. By blending the overlapping positional and rotational references, we enforce strict spatio-temporal consistency, thereby eliminating inference-induced jittering and ensuring seamless transitions between continuous replanning cycles.
    A fundamental prerequisite for stable closed-loop operation is a strictly unified spatial reference. Through data canonicalization during planner training and hardware calibration at controller initialization, both modules are guaranteed to operate within a globally unified world coordinate system, anchored at the robot's initial footprint.

    \begin{figure}[!t]
        \centering
        \includegraphics[width=1\linewidth, height=0.35\linewidth]{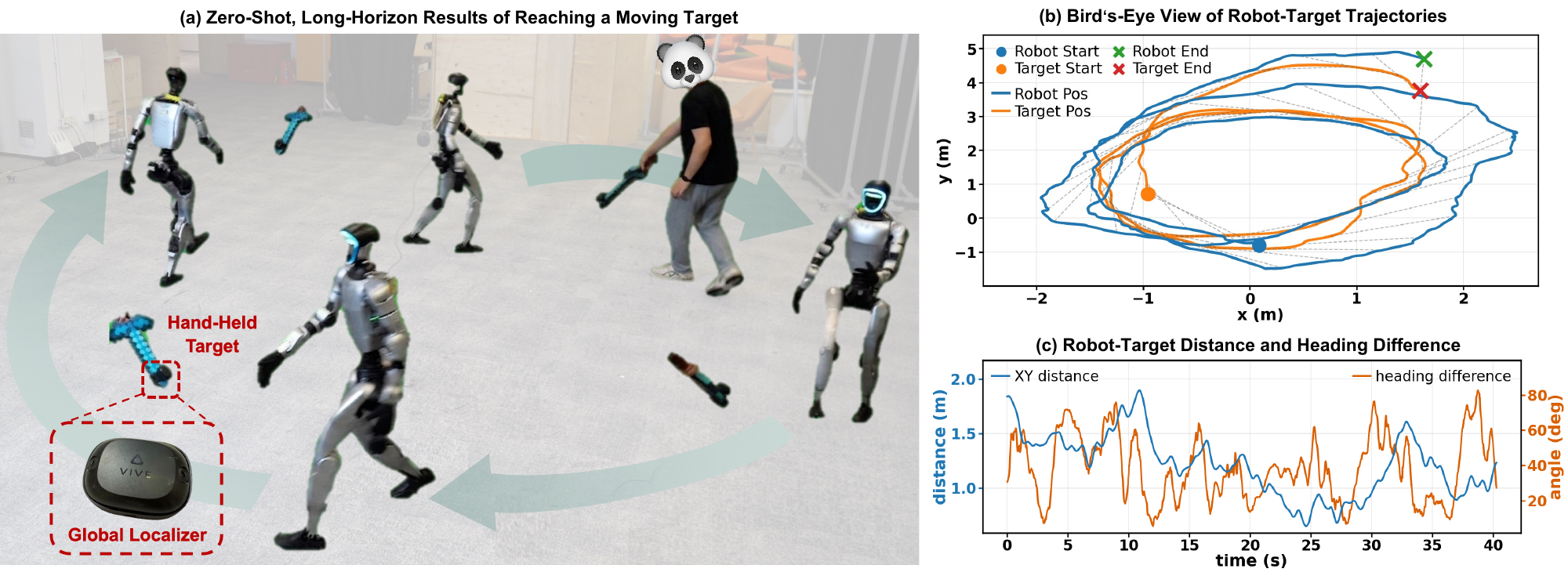}
        \caption{(a) \ours enables long-horizon zero-shot deployment of reaching a moving target. (b) The recorded actual robot-target trajectories show the effectiveness. (c) The curve illustrates that our system reactively plans coordinated whole-body motion and always reaches the moving global localizer.}
        \label{fig:reach}
        \vspace{-15pt}
    \end{figure}

    \paragraphbegin{Zero-Shot Dynamic Target Reaching via Ego-Centric Reset.} 
    To validate the system's dynamic replanning agility in the real world, we design an autonomous task to reach moving targets. 
    % This introduces a twofold challenge. First, empowering a generative planner to perform zero-shot whole-body target reaching without exhaustive real-world task-specific data is fundamentally demanding. Second, while we merely collect a limited training corpus of static reaching~\citep{wang2025physhsi}, deploying this prior against a \textit{moving} target creates a severe distribution shift. This shift arises because the planner has exclusively observed targets that remain static relative to a fixed global origin.
    % % To demonstrate the reactive planning capability of~\ours, we design a task to let a humanoid reach the moving target.
    The core challenge is a severe distribution shift. Our generative planner was trained exclusively on a limited corpus of static reaching data \citep{wang2025physhsi}.
    % To bridge this gap, we propose a continuous ego-centric alignment strategy. At every asynchronous replanning trigger, we mathematically reset the robot's instantaneous root pose as the new global origin, transforming the target's absolute position into this updated egocentric frame. By recursively casting the continuous dynamic tracking problem into a sequence of static-reaching sub-tasks, this mathematically elegant coordinate reset perfectly aligns the on-the-fly observations with the planner's static training prior. While this origin reset mathematically neutralizes the global base velocity, our AR planner seamlessly preserves physical momentum by conditioning on historical proprioception ($\mathbf{x}^k_{pre}$), inherently capturing movement inertia. Consequently, our system leverages these static priors to achieve robust, zero-shot generalization to long-horizon dynamic target reaching.
    To resolve this, we propose a continuous ego-centric alignment strategy. At each replanning step, we reset the robot's instantaneous root pose as the new global origin and transform the target's position into this egocentric frame. This effectively decomposes continuous dynamic tracking into a sequence of static-reaching sub-tasks, perfectly aligning on-the-fly observations with the static training prior. Meanwhile, physical momentum is seamlessly preserved by conditioning our AR planner on historical proprioception. Consequently, the system achieves robust, zero-shot generalization to long-horizon dynamic target reaching. 
    % ($\mathbf{x}^k_{pre}$)

    % Crucially, we achieve this dynamic capability \textit{without} collecting task-specific dynamic tracking data or altering the planner's architecture. Our training corpus derived from PhysHSI only contains sequences of reaching \textit{stable} targets starting from the origin. To bridge this data gap, we introduce a continuous ego-centric alignment strategy. During deployment, at every asynchronous replanning trigger, we mathematically reset the robot's instantaneous root pose as the new global origin. By recursively casting the continuous dynamic tracking problem into a sequence of static-reaching sub-tasks, this mathematically elegant coordinate reset perfectly aligns the on-the-fly observations with the planner's training distribution. While this coordinate reset mathematically neutralizes global base velocity, the physical momentum is seamlessly preserved. By conditioning on historical proprioception ($\mathbf{x}^k_{pre}$), the AR planner inherently captures movement inertia to smoothly extend the transient velocity profile. Consequently, our system achieves zero-shot generalization to long-horizon, dynamic target reaching.

\subsection{Behavior Foundation Model}
\label{method:bfm}
    % We adopt a pretrained BFM~\citep{zeng2026scalebfm} sharing a foundational architecture with SONIC~\citep{luo2025sonic} and earlier BFM designs~\citep{zeng2025behavior}. This universal tracking controller is trained on over 102 million frames in 50FPS of retargeted motion data via PPO~\citep{schulman2017proximal}. The optimization heavily relies on a comprehensive global tracking reward to ensure strict motion fidelity. Extensive domain randomization equips the model with robust adaptability to out-of-distribution state deviations. This inherent resilience perfectly complements our closed-loop planning and control paradigm.

    We employ ScaleBFM~\citep{zeng2026scalebfm} as our low-level controller. ScaleBFM is a large-scale, pretrained humanoid behavior foundation model capable of robust and versatile whole-body control across diverse behavioral specifications. By formulating whole-body motion tracking as a unified proxy task, ScaleBFM is pretrained via PPO~\citep{schulman2017proximal} on a corpus exceeding 100 million retargeted human motion frames. Built upon an expressive and scalable architecture known as the Humanoid Transformer, the model naturally induces structured latent representations of behavioral intentions. Through a carefully designed scaling recipe, ScaleBFM achieves superior control accuracy, robustness, and generalization across diverse scenarios, making it an ideal backbone for our system.
    
    Specifically, we integrate the Whole-Body and Global control modes of ScaleBFM, with root localization provided by an HTC Vive Ultimate Tracker. This configuration enables accurate global motion tracking while maintaining strong resilience to out-of-distribution state deviations, thereby seamlessly complementing our closed-loop planning and control framework.

    \begin{figure}[t]
        \centering
        \includegraphics[width=\linewidth,height=0.45\textheight]{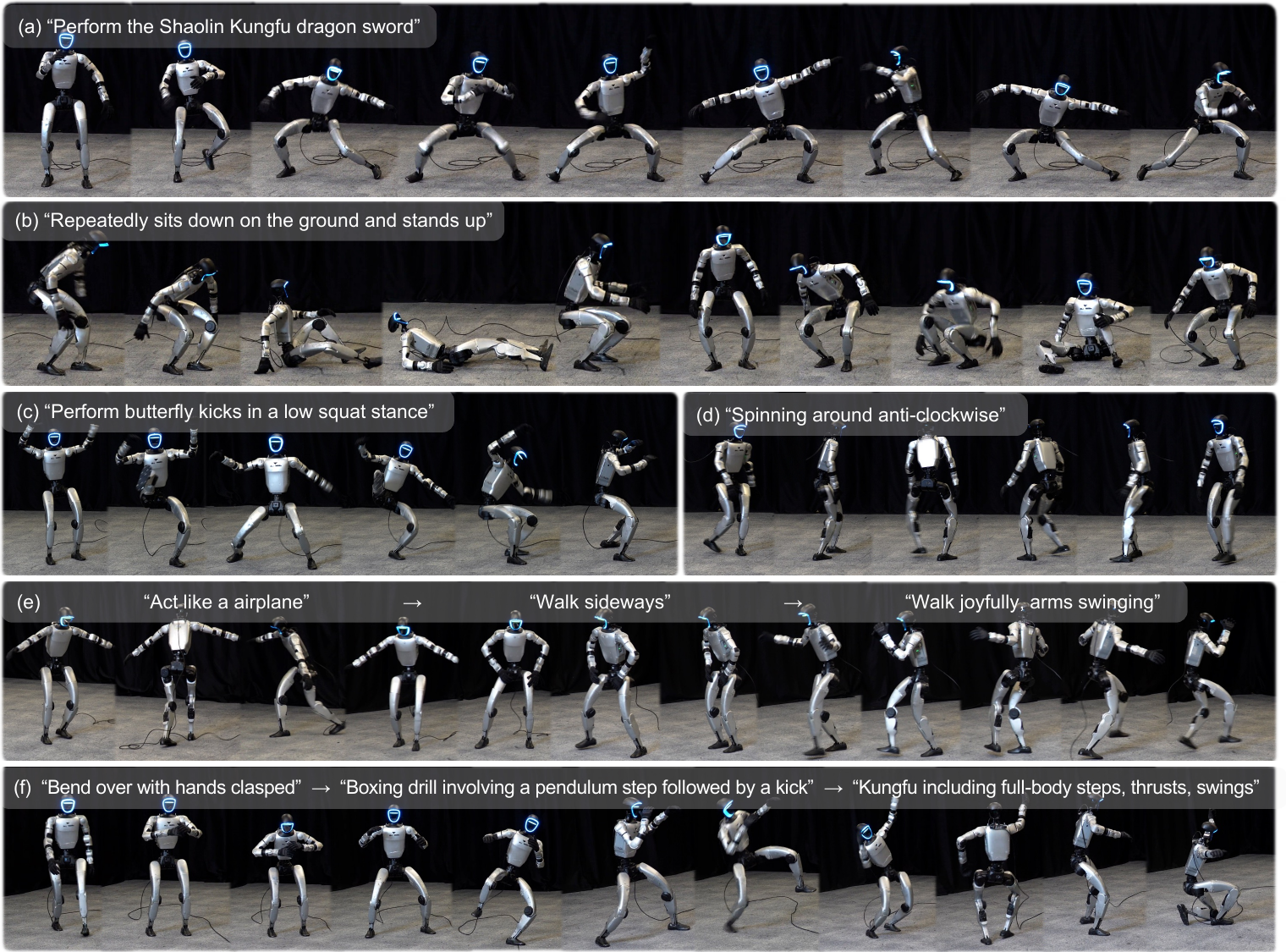}
        \caption{\textbf{Real-world deployment of \ours under text-conditioned and streaming interactive control.} (a)-(d) Diverse maneuvers demonstrate whole-body stability, contact-rich robustness, high-dynamic agility, and high-precision trajectory tracking across diverse skills. (e)-(f) Streaming interactive execution highlights semantic adherence and whole-body coordination under continuous text commands.}
        \label{fig:real}
        \vspace{-10pt}
    \end{figure}

%% file: sections/4_experiments.tex
\vspace{-3pt}
\section{Experiments}
\begin{figure}[!t]
        \centering
        \includegraphics[width=1\linewidth]{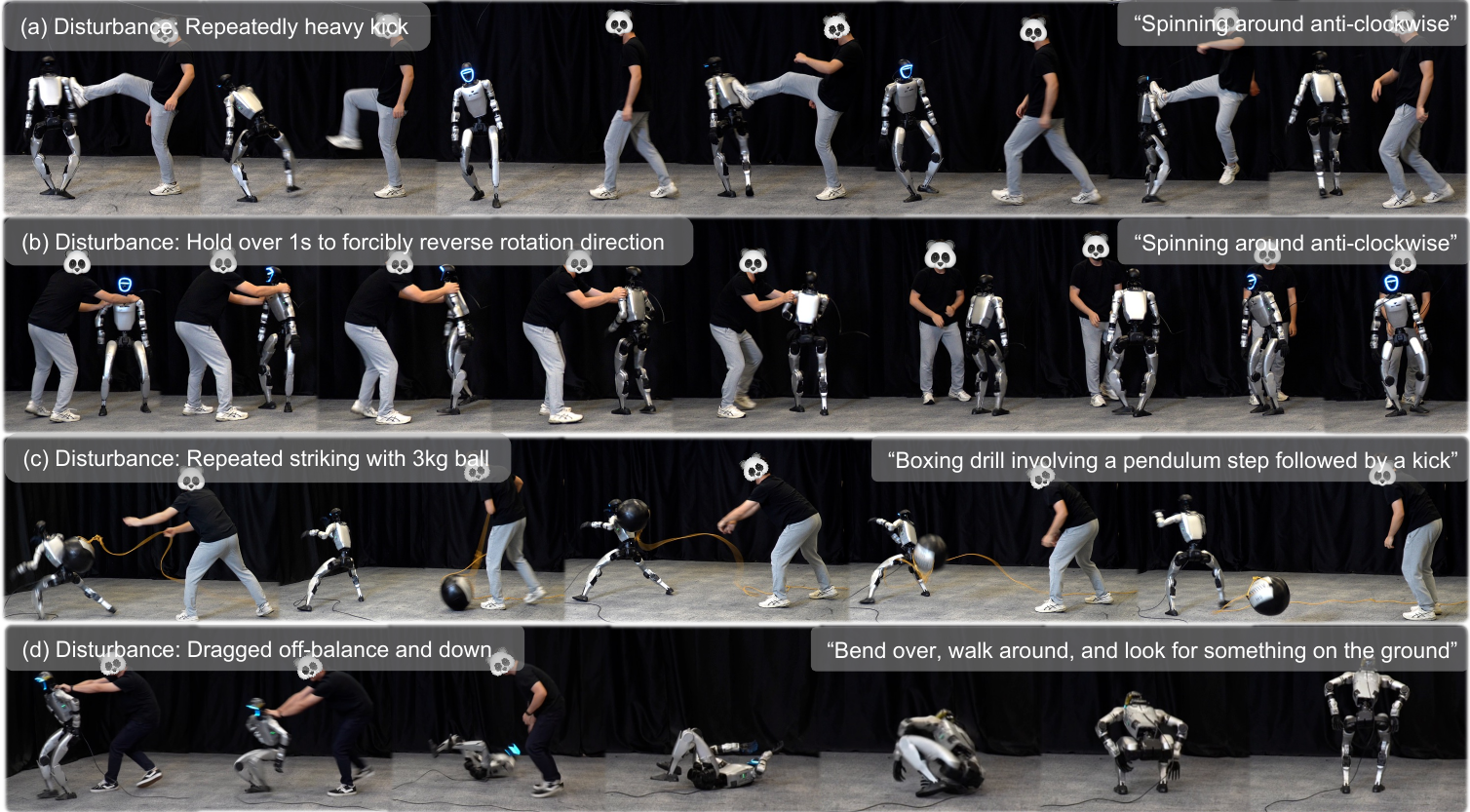}
        \caption{Real-world robustness evaluation under diverse physical perturbations: (a) repeated heavy kicks; (b) holding for over 1s to forcibly reverse the rotation direction; (c) repeated strikes with a 3kg ball; and (d) being dragged off-balance and down.}
        \label{fig:robust}
        \vspace{-16pt}
    \end{figure}

\vspace{-5pt}
    Our experiments address a core system-level question: \emph{\textbf{can a humanoid be operated as a real-time closed-loop planning-control system that continuously replans whole-body motion from multi-modal feedback under real-world disturbances?}} % streaming intent, dynamic targets, and physical feedback
    Rather than treating the planner and controller as isolated modules, we evaluate their integrated closed-loop performance.
    We first demonstrate the framework's robustness and scalability through diverse real-world deployments as shown in Fig.~\ref{fig:reach}, \ref{fig:real}, and~\ref{fig:robust}. 
    Next, we benchmark against baselines and ablate key design choices via sim-to-sim evaluation in Table~\ref{tab:main}. 
    Finally, we profile system latency and runtime strategies in Fig.~\ref{fig:timeline} to validate our real-time performance claims.

    \input{tables/main}

    \input{tables/matched}

\subsection{Experimental Setup}
\label{exp:setup}
\vspace{-5pt}
    \paragraphbegin{Baselines.} We compare our integrated framework against cascaded open- and closed-loop systems, formed by pairing generative planners~\citep{zhao2025dartcontrol,xie2026textop,rempe2026kimodo} retargeting with tracking controllers~\citep{ze2025twist2,luo2025sonic}. For system-level comparison, we pair them to build open-loop and closed-loop pipelines.

    \paragraphbegin{Evaluation \& Metrics.} To evaluate system-level performance under disturbances, we report task success, fall rate, and root-relative mean per-joint position error ($E^r_{MPJPE}$). We introduce replanning smoothness, computed as the average root-relative MPJPE of the first frame across all replanning cycles. We evaluate two execution horizons: open-loop systems execute each offline-generated motion once for its native duration ($1\times$), whereas closed-loop systems continuously replan for an extended horizon of $5\times$ the average motion duration, with survival time capped at 32\,s. Since these horizons are not directly comparable, Table~\ref{tab:matched} further reports duration-matched success rates of open- and closed-loop variants at both horizons. The text-to-motion evaluations in Tables~\ref{tab:main} and \ref{tab:left_table} are conducted on a curated test set comprising 1,000 text-motion pairs sampled from 100STYLE~\citep{mason2022real}, with one episode per pair. Crucially, neither our planner nor the evaluated baselines have been exposed to this test set during training.

    \paragraphbegin{Deployment.}
    All real-world experiments are conducted on a Unitree G1, driven by an off-board workstation with an RTX 4090 GPU. % i9-13900K CPU
    During deployment, the moving target's global pose is captured via an HTC Vive Ultimate Tracker as shown in Fig.~\ref{fig:vive_tracker} and continuously streamed to the planner via Wi-Fi broadcast.
    A 50 Hz low-level controller tracks motions generated by an asynchronous planner, which autoregressively predicts 40-frame chunks from the latest 20-frame observation history. 
    More deployment, training, and evaluation details are provided in the appendix.

\vspace{-6pt}
\subsection{Real-World Deployment}
\label{exp:real}
\vspace{-3pt}
    \paragraphbegin{Zero-Shot Dynamic Target Reaching.}
    % We first evaluate whether~\ours can zero-shot transfer static reaching priors to dynamic long-horizon tracking. 
    Continuous moving targets introduce severe sim-to-real gaps, as varying physical momentum and execution delays easily break open-loop plans. \ours solves this by aggressively replanning in the robot's instantaneous local frame while preserving the autoregressive action prefix to maintain momentum. As shown in Fig.~\ref{fig:reach}, our closed-loop system achieves zero-shot dynamic target reaching, sustaining continuous execution for over 40 seconds. Across 10 real-world trials, the system maintains a robust 90\% success rate without any dynamic-target training data. Ultimately, this demonstrates that our closed-loop architecture can efficiently reuse limited static priors to solve harder, online control problems on physical hardware.

    % \paragraphbegin{Streaming Interactive Control and Task Switching.}
    % We next test whether the robot can execute a continuous stream of text-conditioned whole-body behaviors while receiving new commands online. 
    % % In this setting, the system must not only synthesize diverse motions, but also connect each newly requested behavior from the robot's actual current state. Open-loop motion generation is poorly suited to this regime, since restarting from offline references would introduce discontinuities and ignore accumulated tracking deviations.
    % As shown in Fig.~\ref{fig:real}, in real-world deployment,~\ours supports seamless transitions across long-horizon locomotion, expressive upper-body motions, and highly dynamic whole-body maneuvers. The planner conditions on the latest proprioceptive prefix and the current textual instruction, allowing new commands to overwrite the previous intent without resetting the robot. These results validate that our training strategy for condition switching and proprioceptive grounding translates into practical interactive control on hardware.
    \paragraphbegin{Streaming Interactive Control and Task Switching.}
    Streaming continuous, text-conditioned behaviors introduces a critical challenge: open-loop generation creates discontinuous joint references when new commands arrive. 
    As shown in Fig.~\ref{fig:real} (e-f), in real-world deployment,~\ours supports seamless transitions across long-horizon locomotion, expressive upper-body motions, and highly dynamic whole-body maneuvers. The planner conditions on the latest proprioceptive prefix and the current textual instruction, allowing new commands to overwrite the previous intent without resetting the robot. These results validate that our training strategy for condition switching and proprioceptive grounding translates into practical interactive control on hardware.
    
    \paragraphbegin{Robustness Under Physical Disturbances.}
    % Finally, we evaluate whether feedback-driven replanning improves robustness when the robot is pushed away from the intended motion manifold. Fig.~\ref{fig:robust} summarizes disturbance tests in which the robot is perturbed during online execution. The closed-loop system recovers by conditioning future planning windows on the actually executed state, whereas open-loop execution can only continue tracking references generated before the disturbance.
    % We complement real-world perturbation trials with controlled simulation comparisons against open-loop variants. This allows us to measure survival time, fall rate, tracking error, and task success under repeatable disturbance conditions. The results show that robustness is not merely inherited from the low-level tracker; it emerges from repeatedly closing the loop between the executed state and the next generated reference.
    We finally evaluate system recovery from external physical perturbations. When pushed outside the intended motion manifold, open-loop baselines fail by rigidly tracking obsolete references. In contrast, during real-world deployments (Fig.~\ref{fig:robust}), \ours recovers by dynamically conditioning future planning windows on the forcefully altered physical state. To further quantify this capability, we benchmark the survival time, fall rate, and tracking error in Table~\ref{tab:main}. The results consistently demonstrate superior recovery over open-loop variants. Crucially, this confirms that physical robustness is not merely inherited from the low-level tracking controller, but actively emerges from continuously closing the loop at the planning level.

\vspace{-3pt}
\subsection{Analysis}
\vspace{-5pt}
    
    \paragraphbegin{Sim-to-Sim Evaluation.} As shown in Table~\ref{tab:main}, while open-loop methods and integration baselines such as TextOp+SONIC struggle with physical perturbations, our closed-loop system significantly outperforms all baselines with a 93.1\% success rate, reducing the fall rate to 2.0\% and maintaining smooth motion. The duration-matched results in Table~\ref{tab:matched} isolate the source of this gain: extending execution from $1\times$ to $5\times$ degrades open-loop \ours from 75.2\% to 62.3\%, whereas closing the loop at the same $5\times$ horizon raises success to 93.1\% (+30.8\%). The same trend holds for the retrained TextOp baselines and for our ablation without self-rollout (53.2\% to 70.5\% at $5\times$), confirming that executed-state feedback, rather than the longer horizon, drives the robustness gain.

\begin{figure}[t]
        \centering
        \includegraphics[width=1.0\linewidth, height=0.14\linewidth]{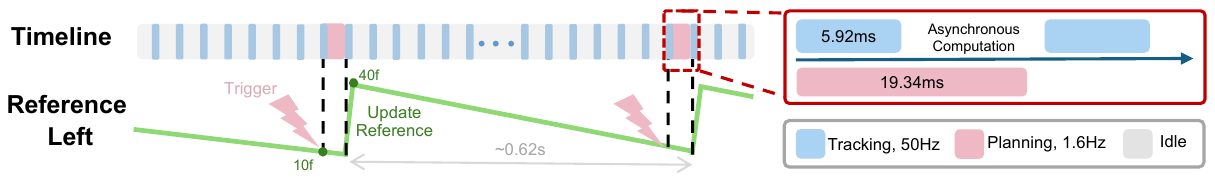}
        \vspace{-10pt}
        \caption{Timeline and latency analysis during real-world deployment.}
        \label{fig:timeline}
        \vspace{-15pt}
    \end{figure}

    \paragraphbegin{Timeline \& Latency Analysis.} Fig.~\ref{fig:timeline} illustrates the runtime behavior of the deployed system. TensorRT acceleration reduces planner inference to 19.3 ms and controller inference to 5.9 ms, allowing the controller to maintain a 50 Hz execution loop while the planner runs asynchronously. Temporal ensembling resolves overlapping prediction chunks and prevents replanning-induced jitter. 
    % These measurements confirm that the proposed architecture is not only algorithmically closed-loop, but also real-time deployable on the physical robot.

%% file: tables/main.tex
% \textbf{FID} & \textbf{R-Precision} 

    % % No external force
    % \begin{table}[!t]
    %     \centering
    %     \setlength{\tabcolsep}{3.5pt}
    %     \renewcommand{\arraystretch}{1.1}   % height
    %     \captionof{table}{The quantitative comparison.}
    %     \begin{tabular}{cccccc}
    %     \hline
    %         \textbf{Methods} & \textbf{Success}$\uparrow$ & \textbf{Fall Rate}$\downarrow$ & \textbf{$E^r_{MPJPE}$}$\downarrow$ & \textbf{Smoothness}$\downarrow$ & \textbf{Survival}$\uparrow$ \\ \hline
    %         % latency
    %         \multicolumn{6}{c}{\textbf{Open-Loop System}} \\ \hline
    %         DART+GMR+SONIC & 73.0\% & 2.0\% & 37.9mm & - &  - \\
    %         TextOp+SONIC & 89.5\% & 10.5\% & 33.5mm & -  & - \\ % 4.5s 
    %         Kimodo+SONIC & 90.2\% & 5.8\% & 36.0mm & - & - \\
    %         \makecell{\textbf{\ours} \\ (Open-Loop)} & 89.9\% & 0.6\% & 30.4mm & - & -\\ \hline % 6.4s
    %         \multicolumn{6}{c}{\textbf{Closed-Loop System}} \\ \hline
    %         TextOp+SONIC & 91.6\% & 6.0\% & 36.7mm & 97.8mm & ~\xiao{tbd} \\
    %         % \makecell{\ours \\ (with Dense Repre.)} & 70.16\% & 37.60mm & & 0.65\% & 31.9s \\
    %         \makecell{\ours \\ (w/o Smoothness Loss)} & 97.1\% & 2.0\% & 33.6mm & 86.7mm & 30.4s \\
    %         \makecell{\ours \\ (Teacher Forcing-Only)} & \% & & & \% & \\
    %         \textbf{\ours} & 100.0\% & 0.0\% & 31.1mm & 82.3mm & 32.0s\\
    %     \hline
    %    \end{tabular}
    %     \label{tab:placeholder}
    % \end{table}
    \begin{table}[!t]
        \centering
        \setlength{\tabcolsep}{3.5pt}
        \renewcommand{\arraystretch}{1.1}   % height
        \captionof{table}{\textbf{Quantitative comparison of open-loop and closed-loop systems under dynamic disturbances.} A random external force of 100N is applied to the humanoid's torso and pelvis for 0.1s during execution. Open-loop systems execute each offline-generated motion once for its native duration ($1\times$), whereas closed-loop systems continuously replan over an extended $5\times$ horizon; the two groups are therefore not duration-matched by design (see Table~\ref{tab:matched} for duration-matched comparisons). Each cell is averaged over 1{,}000 episodes. Hyphens (-) indicate metrics not applicable to open-loop executions.}
        \vspace{2pt}
        \begin{tabular}{cccccc}
        \hline
            \textbf{Methods} & \textbf{Success}$\uparrow$ & \textbf{Fall Rate}$\downarrow$ & \textbf{$E^r_{MPJPE}$}$\downarrow$ & \textbf{Smoothness}$\downarrow$ & \textbf{Survival}$\uparrow$ \\ \hline
            \multicolumn{6}{c}{\textbf{Open-Loop System ($1\times$ native duration)}} \\ \hline
            DART+GMR+SONIC & 51.0\% & 12.5\% & 46.2mm & - &  - \\
            TextOp+SONIC & 64.5\% & 23.5\% & 40.1mm & -  & - \\
            Kimodo+SONIC & 70.4\% & 16.0\% & 41.2mm & - & - \\
            \textbf{\ours} & \textbf{75.2\%} & \textbf{4.3\%} & \textbf{36.5mm}  & - & -\\ \hline
            \multicolumn{6}{c}{\textbf{Closed-Loop System ($5\times$ duration)}} \\ \hline
            TextOp+TWIST2 & 56.1\% & 31.9\% & 51.7mm & 154.6mm & 20.1s \\
            TextOp+SONIC & 76.4\% & 14.7\% & 42.3mm & 128.5mm & 27.8s \\
            \ours with dense repre. & 89.1\% & 2.7\% & 40.3mm & 110.9mm & 29.2s \\
            \ours w/o temp. loss & 83.0\% & 7.5\% & 38.2mm & 118.3mm & 26.9s \\
            \ours w/o self-rollout & 70.5\% & 9.3\% & 41.9mm & 124mm & 26.8s \\
            \textbf{\ours} & \textbf{93.1\%} & \textbf{2.0\%} & \textbf{34.6mm} & \textbf{96.9mm} & \textbf{29.8s}\\
        \hline
        \end{tabular}
        \label{tab:main}
        \vspace{-10pt}
    \end{table}

%% file: tables/matched.tex
\vspace{-5pt}
    \begin{table}[!t]
        \centering
        \setlength{\tabcolsep}{4pt}
        \renewcommand{\arraystretch}{1.1}   % height
        \captionof{table}{\textbf{Duration-matched success rates of open- and closed-loop variants.} All variants are evaluated on the same 100STYLE test set and disturbance protocol as Table~\ref{tab:main}, at both $1\times$ and $5\times$ horizons, isolating the effect of closing the loop from that of the evaluation horizon. *: planner retrained on our training data; TextOp*+ScaleBFM pairs the retrained planner with the same frozen controller as \ours.}
        \vspace{2pt}
        \begin{tabular}{lccc}
        \hline
            \textbf{Methods} & \textbf{Open-Loop ($1\times$)} & \textbf{Open-Loop ($5\times$)} & \textbf{Closed-Loop ($5\times$)} \\ \hline
            TextOp* & 68.4\% & 53.0\% & 76.7\% \\
            TextOp*+ScaleBFM & 70.8\% & 57.6\% & 82.5\% \\
            \ours w/o self-rollout & 73.6\% & 53.2\% & 70.5\% \\
            \textbf{\ours} & \textbf{75.2\%} & \textbf{62.3\%} & \textbf{93.1\%} \\
        \hline
        \end{tabular}
        \label{tab:matched}
        \vspace{-16pt}
    \end{table}

%% file: sections/5_conclusion.tex
\vspace{-2pt}
\section{Conclusion \& Limitations}
\vspace{-4pt}

    We presented \ours, a closed-loop planning-control framework for reactive humanoid control in real-world deployment. We solve cumulative exposure bias through scheduled prefix sampling, reduce physical drift via compact state conditioning, and resolve execution latency by asynchronous replanning with trajectory chunking, while also enabling smoother command transitions via condition dropout and temporal regularization.
    On Unitree G1, our approach achieves 90\% success in zero-shot dynamic goal reaching and sustains over 40 seconds of continuous execution. In sim-to-sim benchmarking under severe perturbations, it attains 93.1\% task success with a 2.0\% fall rate, surpassing cascaded open-loop baselines by 28.6\%. 
    These results highlight the effectiveness of closed-loop generative planning for robust humanoid autonomy.

    % \paragraphbegin{Limitations.}
    % While ~\ours demonstrates strong real-world robustness and scalability, current limiting assumptions introduce potential failure modes.
    % \textit{(1) Humanoid-Object Interaction (HOI):} 
    % % To validate core scalability, we assume humanoid-only motions without precise manipulation. 
    % % Consequently, the system may fail in contact-rich environments where explicit whole-body HOI is mandatory. 
    % The system may fail in contact-rich environments where precise whole-body HOI is mandatory. 
    % Integrating emerging HOI planners to unlock loco-manipulation remains our immediate next frontier.
    % \textit{(2) Extended Sensory Modalities:} 
    % We currently operate under the minimal representative intent-, perception-, and kinematics-level states provide sufficient conditioning for the online motion planner.
    % % While this ensures real-time efficiency, 
    % However, it may lead to failure modes in unconstrained task scenarios that strictly require high-dimensional sensory feedback, such as real-time visual streams or tactile sensing, for precise reactive adjustments. Future work will explore integrating these modalities to address more complex, reactive scenarios.
    % % (3) Multi-Task Capability. In this work, we design dynamic target reaching, interactive streaming control, and perturbation robustness evaluation to show the versatility of our framework.

    \paragraph{Limitations.}
    Despite strong real-world robustness and scalability, \ours relies on two simplifying assumptions. 
    First, it does not explicitly model precise whole-body humanoid-object interaction, and may therefore fail in contact-rich tasks requiring loco-manipulation. 
    Second, the online planner is conditioned only on compact target positions, text commands, and kinematic states, which may be insufficient for tasks requiring high-dimensional vision or tactile feedback. Extending the framework to HOI and richer sensory modalities is an important direction for future work.

% %% Use plainnat to work nicely with natbib. 
% \section*{Acknowledgments}

%% file: sections/6_appendix.tex
\clearpage
\setcounter{page}{1}

\appendix

\section{Additional Results and Analysis}
\subsection{Modular Evaluation}
    We provide a modular evaluation of our system's core components: the motion planner and the tracking controller. The motion planners are evaluated under a standard text-to-motion generation setting, whereas the controllers are assessed under a standard motion tracking setting.

    \paragraphbegin{Metrics.} To evaluate planner quality, we report FID, MM-Dist, and R-Precision@3 (R@3). For controller fidelity, we report the tracking success rate alongside root-relative $E^r_{\mathrm{MPJPE}}$ and absolute mean per-joint position errors $E^a_{\mathrm{MPJPE}}$. An episode is considered a failure if any of the following conditions are met: (a) the humanoid falls (defined as the torso or pelvis height dropping to $\leq$ 30cm); (b) the single-frame $E^r_{\mathrm{MPJPE}} \geq$ 100mm; or (c) the global xy-drift $\geq$ 3m.
    
    \paragraphbegin{Baselines.} For motion planners, we compare against the robot motion generators TextOp~\citep{xie2026textop} and Kimodo~\citep{rempe2026kimodo}, as well as the human motion generator DART~\citep{zhao2025dartcontrol} paired with GMR~\citep{araujo2025retargeting} for retargeting. For tracking controllers, we benchmark against TWIST2~\citep{ze2025twist2} and SONIC~\citep{luo2025sonic}.
    
    \paragraphbegin{Test Set.} 
    The text-to-motion evaluations in Tables~\ref{tab:main} and \ref{tab:left_table} are conducted on a curated test set comprising 1,000 text-motion pairs sampled from 100STYLE~\citep{mason2022real}. Crucially, neither our planner nor the evaluated baselines have been exposed to this test set during training. Similarly, the tracking evaluation (Table~\ref{tab:param_control}) utilizes this 100STYLE test set. Both the baseline controllers and our BFM are evaluated on this dataset without any prior training on 100STYLE.

    \vspace{-5pt}
    \input{tables/modules}

    % \subsection{Qualitative Comparison}
    % We also provide the sim-to-sim evaluation to show the robustness of our closed-loop planning-control system~\ours compared with open-loop baselines.~\xiao{TODO: sim2sim robust}

    \subsection{Onboard Execution and Inference Latency}
        
    \ours supports fully onboard execution, with the planner and controller meeting their required 1.6-Hz and 50-Hz rates respectively, as shown in Table~\ref{tab:latency}.
    
      \begin{table}[!h]
      \centering
      \footnotesize
      \setlength{\tabcolsep}{5pt}
      \renewcommand{\arraystretch}{1.2}
      \vspace{-9pt}
      \captionof{table}{Inference latency averaged over 10,000 runs.}
      % \vspace{-6pt}
      \label{tab:latency}
      \begin{tabular}{@{}lcccc@{}}
      \toprule
      \multirow{2}{*}{\textbf{Device}}
      & \multicolumn{2}{c}{\textbf{Planner} (required:\ 1.6\,Hz)}
      & \multicolumn{2}{c}{\textbf{Controller} (required:\ 50\,Hz)} \\
      \cmidrule(lr){2-3}\cmidrule(l){4-5}
      & TensorRT & PyTorch & TensorRT & PyTorch \\
      \midrule
      \textbf{Jetson Orin} & 196.2ms/5Hz & 300.3ms/3Hz & 12.8ms/78Hz & 62.7ms/16Hz \\
    
      \textbf{Jetson Thor} & 38.2ms/26Hz & 50.8ms/20Hz & 3.3ms/306Hz
      & 8.1ms/124Hz \\
    
      % \textbf{RTX 4090D} & 19.3ms / 52Hz &  & 5.9ms / 169Hz &  \\
      \textbf{RTX 5090D} & 13.2ms/75Hz & 23.3ms/43Hz & 2.0ms/495Hz & 5.5ms/182Hz \\
      \bottomrule
      \end{tabular}
      % \vspace{-10pt}
      \end{table}

\subsection{Analysis for failure cases}
    \noindent \textbf{1) Global localization.}
    Across $>$100 general trials, pose outages exceeding $1\,\mathrm{s}$ occurred in $<5\%$ of trials. In dual-tracker moving-target trials, nearly all observed failures stemmed from localization loss when either VIVE tracker left the $15\,\mathrm{m}\times15\,\mathrm{m}$ reliably mapped area, rather than unrecoverable policy errors. 
    Reduced gait smoothness primarily results from communication-induced temporal misalignment between the two VIVE trackers, exacerbated by the high-frequency updates required in this zero-shot.
    
    \noindent \textbf{2) Planner.} 
    With valid localization, our planner handles fast targets, sharp omnidirectional turns, and sustains long-term rollout. Observed failures mainly arise after prolonged execution ($\geq45\,\mathrm{s}$) with multiple skill switches, occasionally degenerating to a neutral pose or transition jitter due to accumulated rollout error or conflicting contexts. 

\vspace{-5pt}
\section{Implementation Details}
\label{appendix:details}
    % \paragraphbegin{Lightweight Architecture.}
    % To ensure the real-time inference frequency, our planner adopts a lightweight architecture employing an 8-layer transformer decoder, following~\citep{tevetclosd,chen2024taming}.

\subsection{Evaluation Details}
    % It is important to emphasize that the experimental results presented in the main text \textbf{do not rely on any ground-truth reference motions}. Also, we evaluate the pretrained BFM in a completely frozen state, without any task-specific fine-tuning. Consequently, the controller's ability to seamlessly track these generated motions serves as a rigorous demonstration of its zero-shot generalization capabilities.
    
    It is important to emphasize that all evaluated motions in the main paper are \textbf{generated via real-time online planning, without utilizing any pre-existing reference motions}. Also, we evaluate the pretrained BFM directly in a completely frozen state without any form of downstream fine-tuning. Consequently, the controller's ability to seamlessly track these dynamically planned trajectories in real time serves as a rigorous demonstration of its robust zero-shot generalization capabilities.
    
\subsection{Model Architecture}
    \paragraphbegin{Planner.} To ensure the real-time inference frequency, our planner adopts a lightweight architecture employing an 8-layer transformer decoder~\citep{vaswani2017attention}, following~\citep{tevetclosd,chen2024taming}.

    \paragraphbegin{Controller.}
    We introduce the used BFM~\citep{zeng2026scalebfm} here.
    To provide rich temporal information for behavior learning, the controller utilizes a Transformer-based architecture that processes sequential historical context and future control specifications within a finite temporal window. 
    
    Specifically, given a history length $L$, the proprioceptive states $s^P$ and executed actions $a$ are projected into embeddings using separate, modality-specific tokenizers with parameters shared across matching modalities. These historical embeddings are sequence-interleaved and appended with a learnable query token $e$ at the end to formulate the complete context sequence $\mathcal{Z}_t^p = (z_{t-L+1}^{s^p}, z_{t-L+1}^{a}, \dots, z_{t-1}^{s^p}, z_{t-1}^{a}, z_{t}^{s^p}, e)$. Concurrently, future goal states covering a window of $N$ future frames are paired with their relative temporal offsets and encoded into task tokens $\mathcal{Z}_t^g = (z_1^{s^g}, z_2^{s^g}, \dots, z_N^{s^g})$.
    
    The backbone consists of multiple transformer layers, each incorporating self-attention, cross-attention, a feed-forward network (FFN)~\citep{vaswani2017attention} implemented via SwiGLU~\citep{shazeer2020glu}, and Root Mean Square Normalization (RMSNorm)~\citep{zhang2019root}. Within the self-attention block, causal masking prevents the context tokens from attending to the query token, while the query token can attend to all context tokens to aggregate historical trajectory details. Rotary Position Embedding (RoPE)~\citep{su2024roformer} is applied to encode positional relations. Cross-attention blocks are then leveraged to inject the task tokens $\mathcal{Z}_t^g$ into the context sequence, conditioning the behavioral generation on flexible control specifications. 
    Notably, an RMSNorm module maps the goal embeddings onto a continuous, bounded sphere, encouraging the natural emergence of structured behavioral representations. Finally, the hidden representation corresponding to the query token $\hat{e}$ is fed into a projection head consisting of a single linear layer to predict actions or values.

\input{tables/parameter}

\subsection{Global Localization}
\begin{figure}[h]
    \centering
    \includegraphics[width=0.9\textwidth]{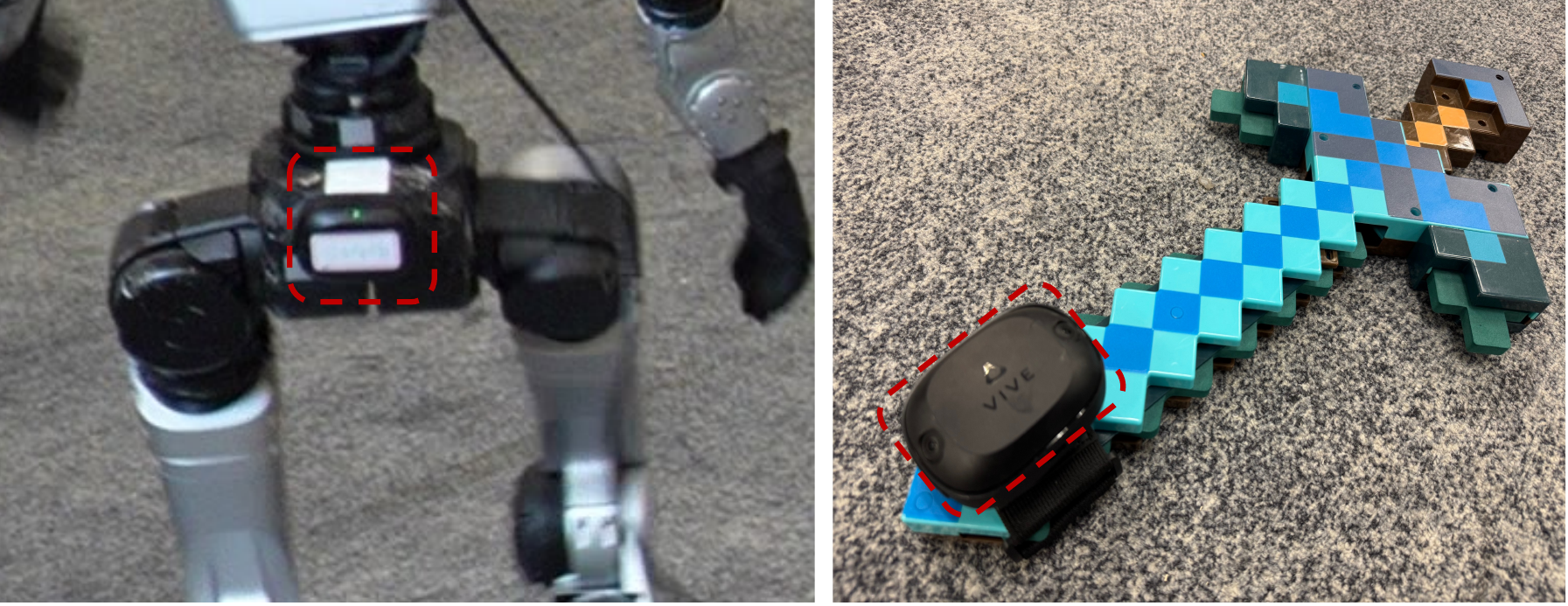}
    \caption{We utilize HTC VIVE Ultimate Trackers for global localization. During deployment, one tracker is mounted on the back of the robot’s pelvis, and another is attached to a handheld toy sword.}
    \label{fig:vive_tracker}
\end{figure}

    In the dynamic target reaching experiments, we employ a control paradigm that combines an online planner with global tracking. This requires precise, real-time global localization for both the robot and the handheld moving target. We utilize HTC VIVE Ultimate Trackers as shown in Fig.~\ref{fig:vive_tracker} as our underlying localization hardware. Prior to deployment, we perform a global mapping of the workspace and calibrate the origin of the world coordinate system. Subsequently, one tracker is rigidly mounted to the rear of the robot's pelvis using a custom connector to capture the global base state of the humanoid. Concurrently, a second tracker is securely strapped to the human operator's hand to track the spatial trajectory of the moving target. During online execution, a dedicated Windows laptop receives and processes the global localization data. This node then streams the coordinate information to the primary deployment host via TCP over a local area network (LAN) at 100~Hz, ensuring high-bandwidth and low-latency responses for the real-time control loop.

    We note that external localization is task-dependent rather than required by the low-level policy. Our framework is fully flexible to modes including local (w/o global localization) and global tracking.

\subsection{Key Hyperparameters}
\label{app:param}
    The key hyperparameters of our planner and controller are shown in Table~\ref{tab:param_planner} and Table~\ref{tab:param_control}. Our BFM is trained on IsaacLab~\citep{mittal2025isaaclab} and optimized by PPO~\citep{schulman2017proximal}.
    % implementation from Stable-Baselines3~\citep{raffin2021stable}, which is developed in PyTorch~\citep{paszke2019pytorch}. 

\subsection{Additional Details}
    All real-world experiments are conducted on a Unitree G1 humanoid. The low-level controller runs at 50 Hz, while the motion planner asynchronously generates whenever the execution buffer approaches depletion. At each execution, the motion planner synthesizes 40-frame reference chunks conditioned on the latest 20-frame actual robot trajectories. 
    The planner is trained on a single RTX 4090 within 48 hours. The BFM is pretrained on 64 GPUs for about 10 days using over 102 million frames in 50 FPS of diverse motion data, delivering unprecedented global tracking precision and zero-shot sim-to-real transfer capabilities.
    The deployment computer is equipped with an Intel i9-13900K CPU and an NVIDIA RTX 4090 GPU.

\section{Planner Training Data}
% \section{Training Data}
    % \paragraphbegin{Planner Training Data.}
    To endow the generative planner with physically grounded and highly dynamic capabilities, we construct a comprehensive text-motion dataset by aggregating multiple high-fidelity sources. Our foundational data comprises diverse human motions from AMASS paired with HumanML3D text annotations~\citep{guo2022generating}, supplemented by the large-scale Humanoid-Union dataset~\citep{wei2025unveiling} to explicitly incorporate highly dynamic, robot-specific maneuvers. 
    
    To ensure high-quality supervision, we implement a rigorous data curation pipeline. On the language side, we filter typos and unify synonyms to standardize the textual conditions. On the motion side, we perform kinematic canonicalization and explicitly conduct physical corrections within the PyBullet simulator. This critical step eliminates physically infeasible artifacts from the raw data, ensuring that all reference trajectories are dynamically viable for the humanoid. 
    
    Furthermore, to specifically enhance the planner's capability to reach targets, we synthesize 10,000 task-specific trajectories using a pre-trained PhysHSI policy. By deliberately truncating these reach-carry rollouts to retain only the effective "reach" phases, we inject high-quality manipulation priors into the generation space. Consequently, our finalized corpus yields $\text{37.14}$ hours ($\text{8,021,854}$ frames) of foundational motion with context labels. The motion corpus encompasses highly dynamic maneuvers, whole-body coordination, global locomotion, and daily activities, and $\text{10.15}$ hours ($\text{2,192,001}$ frames) of targeted reaching data. This rigorous curation ensures a robust, semantically aligned, and physically reliable training foundation.
    % 37.14 hours (totally 30,660 episodes, 8,021,854 frames)

    % % New Version
    % Raw motion datasets often contain physically infeasible artifacts that destabilize closed-loop controllers. To establish a robust, physically grounded training corpus, we aggregate data from AMASS~\citep{mahmood2019amass}-HumanML3D~\citep{guo2022generating} and 100STYLE~\citep{mason2022real}\&Kungfu~\citep{lin2023motion} datasets organized by Humanoid-Union~\citep{wei2025unveiling} through a rigorous three-step curation pipeline. First, we perform kinematic canonicalization and explicit physical corrections within the PyBullet~\citep{coumans2016pybullet} simulator to systematically eliminate physically implausible motions. Second, we enforce semantic standardization on the corresponding HumanML3D text annotations by filtering typographical noise and unifying synonyms. Finally, for task-specific augmentation, we synthesize 10,000 ``reach-carry" trajectories using a pre-trained PhysHSI~\citep{wang2025physhsi} policy, strictly truncating them to retain only the effective reaching phases. Ultimately, this curated dataset provides 37.14 hours (total 30,660 episodes, 8,021,854 frames) of motion data with text annotations or target positions. The whole dataset spans highly dynamic maneuvers, whole-body coordination, and global locomotion.

    % \paragraphbegin{Controller Training Data}.

    \begin{table}[t]
    \centering
    \caption{The statistics of the retargeted motion datasets for planner training.}
    \label{tab:dataset_stats}
    \begin{tabular}{lcccc}
        \toprule
        \textbf{Dataset} & \textbf{\#Motions} & \textbf{\#Frames} & \textbf{Avg.\ Length (frames)} & \textbf{Duration @60\,fps} \\
        \midrule
        AMASS-HumanML3D & 11{,}424 & 1{,}156{,}021 & 101.2 & $\approx$5.35\,h  \\
        100STYLE        &  8{,}100 & 4{,}055{,}978 & 500.7 & $\approx$18.78\,h \\
        Kungfu          &  1{,}032 &   617{,}854   & 598.7 & $\approx$2.86\,h  \\
        PhysHSI-Reach   &  9{,}994 & 2{,}192{,}001 & 219.3 & $\approx$10.15\,h \\
        \midrule
        \textbf{Total}  & \textbf{30{,}550} & \textbf{8{,}021{,}854} & \textbf{262.6} & $\approx$\textbf{37.14\,h} \\
        \bottomrule
    \end{tabular}
    \end{table}

%% file: tables/modules.tex
\begin{table}[!h]
    \centering
    
    % ======= Left =======
    \begin{minipage}{0.48\textwidth}
        \centering
        \setlength{\tabcolsep}{3pt}
        \renewcommand{\arraystretch}{1.1}
        \caption{The evaluation of motion planners.}
        \label{tab:left_table}
        \begin{tabular}{cccc}
            \hline
            % \textbf{Methods} & \textbf{FID}$\downarrow$ & \textbf{MM-Dist}$\downarrow$ & \textbf{R@3}$\uparrow$\\ \hline
            % DART+GMR & 39.24 & 8.66 & 0.11 \\
            % TextOp & 15.57 & 7.39 & 0.19 \\
            % Kimodo & 4.83 & 5.79 & 0.28 \\
            % \textbf{Ours} & 0.13 & 3.79 & 0.55 \\ \hline
            \textbf{Methods} & \textbf{FID}$\downarrow$ & \textbf{MM-Dist}$\downarrow$ & \textbf{R@3}$\uparrow$\\ \hline
            DART+GMR & 39.62 & 8.27 & 0.10 \\
            TextOp & 18.20 & 7.11 & 0.15 \\
            Kimodo & 3.83 & 5.25 & 0.35 \\
            \textbf{Ours} & \textbf{2.10} & \textbf{4.69} & \textbf{0.45} \\ \hline
        \end{tabular}
    \end{minipage}
    \hfill
    % ======= Right =======
    \begin{minipage}{0.48\textwidth}
        \centering
        \setlength{\tabcolsep}{3pt}
        \renewcommand{\arraystretch}{1.1}
        \caption{The evaluation of tracking controllers.}
        \label{tab:right_table}
        \begin{tabular}{cccc}
            \hline
            \textbf{Methods} & \textbf{Suc.}$\uparrow$ & \textbf{$E^r_{MPJPE}$}$\downarrow$ & \textbf{$E^a_{MPJPE}$}$\downarrow$ \\ \hline
            TWIST2~\citep{ze2025twist2} & 22.3\% & 94.0mm & 858.6mm \\
            SONIC~\citep{luo2025sonic}  & 71.0\% & 42.7mm & 1171.2mm \\
            \textbf{Ours-Local} & 82.4\%  &  36.6mm & 802.4mm \\
            \textbf{Ours-Global} & \textbf{94.1\%} & \textbf{29.2mm} & \textbf{97.5mm} \\ \hline
        \end{tabular}
    \end{minipage}
    
\end{table}

%% file: tables/parameter.tex
\begin{table}[!t]
    \centering
    
    % ======= Left =======
    \begin{minipage}{0.48\textwidth}
    \centering
    \setlength{\tabcolsep}{5pt}
    \renewcommand{\arraystretch}{1.1}
    \caption{The key hyperparameters for our planner.}
    \label{tab:param_planner}
        \begin{tabular}{ll}
        \toprule 
        \textbf{Term} & \textbf{Value} \\
        \midrule 
        Optimizer & Adam \\
        Optimization batch size & 128 \\
        Learning rate & 1.0e-4 \\
        Training Iterations & 2500 \\
        % Max Episode length \todo & 100 \\
        Training Environments & 32 \\
        N steps & 512 \\
        N epochs & 4 \\
        Buffer size & 30 \\
        Value coefficient & 0.8  \\ 
        Entropy coefficient & 0.01  \\ 
        Discount factor $\gamma$ & 0.99  \\ 
        GAE $\tau$ & 0.99  \\ 
        PPO clipping & 0.2  \\ 
        \bottomrule
        \end{tabular}
    \end{minipage}
    \hfill
    % ======= Right =======
    \begin{minipage}{0.48\textwidth}
    \centering
    \setlength{\tabcolsep}{5pt}
    \renewcommand{\arraystretch}{1.1}
    \caption{The key hyperparameters for our controller.}
    \label{tab:param_control}
        \begin{tabular}{ll}
        \toprule 
        \textbf{Term} & \textbf{Value} \\
        \midrule 
        % Optimizer & Adam \\
        % Optimization batch size & 128 \\
        % Learning rate & 0.0001 \\
        % Training Iterations & 2500 \\
        % % Max Episode length \todo & 100 \\
        % Training Environments & 32 \\
        % N steps & 512 \\
        % N epochs & 4 \\
        % Buffer size & 30 \\
        % Value coefficient & 0.8  \\ 
        % Entropy coefficient & 0.01  \\ 
        % Discount factor $\gamma$ & 0.99  \\ 
        % GAE $\tau$ & 0.99  \\ 
        % PPO clipping & 0.2  \\ 

        Optimizer & Adam \\
        % Optimization batch size & ? \\
        Actor learning rate & 2.0e-5 \\
        Critic learning rate & 1.0e-3 \\
        % Training Iterations & 22200 \\
        % Max Episode length & 100 \\
        Training Environments & 8192 \\
        % N GPUs & 64 \\
        Mini-batch number & 16 \\
        Learning epochs & 2 \\
        Rollout steps & 64 \\
        % Buffer size & ? \\
        Value coefficient & 1.0 \\ 
        Entropy coefficient & 5.0e-3 \\ 
        Discount factor $\gamma$ & 0.99 \\ 
        GAE $\lambda$ & 0.95  \\ 
        PPO clipping & 0.2  \\ 

        % Architecture Parameters
        % Future window size ($N$) & 6 \\
        % Actor future temporal offsets & [1~5, random(6,33)] \\
        % Critic future temporal offsets & [1, 2, 4, 8, 16, 32] \\
        % Adaptive Sampling Parameters
        % Evaluation interval $T_{eval}$ & 200 epochs \\
        Sampling discount factor $\gamma$ & 0.999 \\
        % Adaptive sampling min weight ($w_{min}$) & 0.03 \\
        % Adaptive sampling max weight ($w_{max}$) & 1.0 \\
        % Deviation / Termination Constraints
        % Termination deviation threshold & 0.5 m \\
        % Control Constraints
        PD controller frequency & 200 Hz \\
        BFM inference frequency & 50 Hz \\
    
        \bottomrule
        \end{tabular}
    \end{minipage}
\end{table}